\documentclass{article}
\usepackage{spconf,amsmath,graphicx}

\usepackage{amssymb}
\usepackage{bbm}
\usepackage{threeparttable}
\usepackage{booktabs}
\usepackage{setspace}
\usepackage{hyperref}
\usepackage{fancyhdr}

\title{LEARNING LIGHTWEIGHT PEDESTRIAN DETECTOR WITH HIERARCHICAL KNOWLEDGE DISTILLATION}
\name{Rui Chen, Haizhou Ai, Chong Shang, Long Chen, Zijie Zhuang
\thanks{This work was supported by the National Science Foundation of China (Project Number 61521002).}
}
\address{
  Beijing National Research Center for Information Science and Technology (BNRist), \\ 
  Department of Computer Science and Technology, 
  Tsinghua University, 
  Beijing, 
  China, 
  100084. \\ 
  \{chenr18, shang-c13, l-chen16, zhuangzj15\}@mails.tsinghua.edu.cn, ahz@mail.tsinghua.edu.cn
  } 
\begin{document}
%
\maketitle
\renewcommand{\footnotesize}{\scriptsize}
\thispagestyle{fancy}
\fancyhead{}
\lhead{}
\lfoot{\footnotesize{Copyright 2019 IEEE. Published in the IEEE 2019 International Conference on Image Processing (ICIP 2019), scheduled for 22-25 September 2019 in Taipei, Taiwan. Personal use of this material is permitted. However, permission to reprint/republish this material for advertising or promotional purposes or for creating new collective works for resale or redistribution to servers or lists, or to reuse any copyrighted component of this work in other works, must be obtained from the IEEE. Contact: Manager, Copyrights and Permissions / IEEE Service Center / 445 Hoes Lane / P.O. Box 1331 / Piscataway, NJ 08855-1331, USA. Telephone: + Intl. 908-562-3966.}}
\cfoot{}
\rfoot{}
\begin{abstract}
It remains very challenging to build a pedestrian detection system for real world applications, which demand for both accuracy and speed. 
This work presents a novel hierarchical knowledge distillation framework to learn a lightweight pedestrian detector, which significantly reduces the computational cost and still holds the high accuracy at the same time. 
Following the `teacher--student' diagram that a stronger, deeper neural network can teach a lightweight network to learn better representations, we explore multiple knowledge distillation architectures and reframe this approach as a unified, hierarchical distillation framework. 
In particular, the proposed distillation is performed at multiple hierarchies, multiple stages in a modern detector, which empowers the student detector to learn both low-level details and high-level abstractions simultaneously. 
Experiment result shows that a student model trained by our framework, with 6 times compression in number of parameters, still achieves competitive performance as the teacher model on the widely used pedestrian detection benchmark. 
\end{abstract}
\section{Introduction}
\label{sec:intro}
\begin{figure*}
  \begin{center}
  \includegraphics[width=0.90\linewidth]{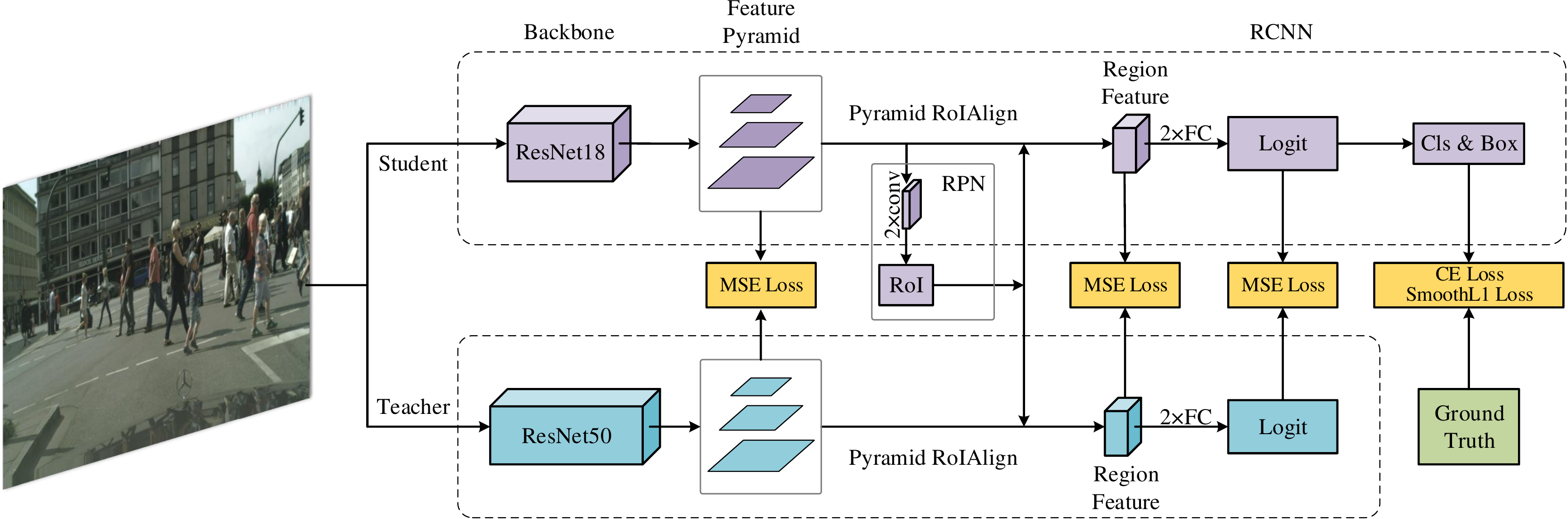}
  \end{center}
  \setlength{\belowcaptionskip}{-10pt} 
  \vspace{-2mm}
  \caption{
    The proposed framework. 
    The stronger ResNet50 detector teaches the ResNet18 to obtain a better representation. 
    Multiple distillations, \textit{i.e.} MSE Loss, are performed in the pipeline. 
    Note that the RPN Loss is omitted in the figure.}
  \vspace{-4mm}
  \label{fig:architecture}
\end{figure*}
Pedestrian detection is a long-standing topic in computer vision. 
It aims to locate and classify pedestrians 
of various sizes and aspect ratios. 
It plays a crucial role in many applications, 
ranging from video surveillance to autonomous driving, 
which usually demand for both fast and accurate detection. 

Recent years have seen a tremendous increase in the accuracy of pedestrian detection, 
relying on deep convolutional neural networks (CNNs) \cite{krizhevsky2012imagenet, Simonyan15, he2016deep}. 
CNN-based object detectors \cite{girshick2014rich, girshick2015fast, ren2015faster, lin2017feature}, such as Faster R-CNN, 
have become the mainstream approaches for object detection. 
With increasingly deeper architecture, these models achieve better performance, which are usually associated with growing computational expense. 
Consequently, 
the speed-accuracy trade-off makes it difficult for these cumbersome detectors to be applied in real-world scenarios. 
In this work, 
we aim at learning a pedestrian detector with both faster speed and satisfied accuracy through hierarchical knowledge distillation. 

Knowledge distillation aims to improve a lightweight model's performance 
by learning from a well-trained but cumbersome model \cite{hinton2015distilling}. 
It usually formulates a teacher-student architecture \cite{romero2014fitnets}, 
which treats the lightweight model as the student, and the cumbersome model as the teacher. 
At first, distillation technique was applied in classification tasks, 
but recent works show great potentials of knowledge distillation in detection tasks \cite{wei2018quantization, Li_2017_CVPR, chen2017learning}. 
Li \emph{et al.}~\cite{Li_2017_CVPR} propose a feature mimicking framework for training efficient object detectors 
to relief the detector training pipeline from ImageNet pre-training. 
They apply distillation technique by adding a supervision to high-level features, 
which helps the small network better learn object representations during training stage.  
Chen \emph{et al.}~\cite{chen2017learning} use the final output of the teacher's region proposal network 
and region classification network as the distillation targets, 
and also adopt a intermediate supervision to improve the student's performance. 

However, these works have several shortcomings. 
Only high-level features are considered to perform distillation in \cite{Li_2017_CVPR}, 
which loses much spatial information. 
Thus their model can not easily detect pedestrians with severe scale variations. 
An intermediate supervision is also considered as a distillation target in \cite{chen2017learning}, 
but it turns out that their student detectors still are lack of supervisions from multiple levels, 
which are crucial to knowledge distillation. 

In this paper, 
we propose a unified hierarchical knowledge distillation framework for pedestrian detection task. 
Firstly, 
instead of adding a supervision to the final feature map, 
we perform knowledge distillation in multiple layers, 
ranging from low-level local details to high-level semantic abstractions, 
referred as pyramid distillation. 
Secondly, we perform pyramid RoIAlign to extract features from multi-level feature maps 
followed with a concatenation operation, 
and apply distillation on all levels, 
which enables the student to see more feature levels with wider receptive field. 
The intuition behind is that 
a student should also learn spatial details as well as high level representations 
to become a comprehensive pedestrian detector. 
Additionally, 
multiple level distillation can server as intermediate supervisions to 
improve the gradient flow during backpropagation.
Finally, 
the distillation is also performed on the logit features following \cite{chen2017learning}, 
with which our approach forms a comprehensive, unified distillation framework. 
Our contributions are three folds: 
\vspace{-3mm}
\begin{enumerate}
  \item We propose an unified knowledge distillation framework, 
  which combines multiple intermediate supervisions for more efficient learning strategy. 
  And existing works, such as \cite{Li_2017_CVPR} and \cite{chen2017learning}, 
  can be viewed as special cases of our proposed framework. 
  \vspace{-3mm}
  \item We propose a hierarchical distillation technique, 
  which adds supervisions to multi-level feature maps, 
  for learning better object representations with both 
  strong semantic abstractions and precise spatial responses. 
  \vspace{-3mm}
  \item We train a lightweight pedestrian detector through our unified, hierarchical distillation framework, 
  which achieves competitive performance on the widely used 
  Caltech \cite{Dollar2012PAMI} pedestrian detection benchmark. 
\end{enumerate}
\section{Related Work}
\label{sec:format}
%
This section briefly discusses previous works that are most related to the proposed framework in object detection and knowledge distillation. 

\textbf{CNNs for Detection.} 
Recently, CNN-based object detectors \cite{girshick2014rich, girshick2015fast} have become the mainstream for detection tasks. 
Regular CNN detectors can be categorized into single-stage \cite{redmon2016you, liu2016ssd, lin2018focal} 
and two-stage \cite{ren2015faster, lin2017feature, shang2018zoomnet} detectors. 
Faster R-CNN \cite{ren2015faster} is a typical two-stage detector, 
which generates region proposals in the first stage, 
and classifies the proposals in the second stage. 
Feature Pyramid Network (FPN) \cite{lin2017feature} 
proposes a top-down architecture with lateral connections 
for building high-level semantic feature maps at all scales. 

\textbf{Knowledge Distillation in Detection.} 
Knowledge distillation aims to improve a lightweight model's performance 
by learning from a well-trained but cumbersome model \cite{hinton2015distilling, romero2014fitnets}. 
Li \emph{et al.}~\cite{Li_2017_CVPR} propose a feature mimicking framework based on high-level feature distillation, 
to train efficient detectors without ImageNet pre-training. 
Wei \emph{et al.}~\cite{wei2018quantization} propose to combine feature mimicking with quantization technique, 
which helps student network to better match the feature maps of teacher network for more efficient feature distillation. 
Chen \emph{et al.}~\cite{chen2017learning} design a distillation framework for detection 
based on final output distillation and single intermediate supervision. 
\section{Method}
\label{sec:pagestyle}
In this section, 
we first present an overview of our framework, 
and then introduce our hierarchical knowledge distillation, 
which are achieved with multiple supervisions:  
Firstly, we perform distillation on multiple feature levels in the pyramid, 
referred as Pyramid Distillation (PD). 
Secondly, we perform another distillation on the output proposals, 
referred as Region Distillation (RD), enabling the student to focus on the positive regions. 
Finally, we add the final distillation at the very end of the detector, 
referred as Logit Distillation (LD).
\subsection{Framework overview} 
%
Our detection framework is built on FPN, 
which introduces a top-down connection to join different levels 
($C_2$ -- $C_5$) for constructing a pyramid of deep features ($P_2$ -- $P_5$) \cite{lin2017feature}. 
Then a region proposal network (RPN) \cite{ren2015faster} is adopted across 
all pyramid features ($P_2$ -- $P_5$) to generate an over-complete set of proposals. 
Following the RPN, we crop features according to the proposals, 
then feed them to a regional classifier to get the final detection results (class labels and boxes). 
In the training stage, we adopt cross entropy loss for 
classification and $smooth_{L_1}$ loss for box regression \cite{girshick2015fast}. 
And the RPN is also trained as Ren \emph{et al.}~\cite{ren2015faster}. 

In this work, we adopt ResNet18 as a student model and ResNet50 as a teacher model. 
In training stage, as illustrated in Fig. 1, 
the input images are fed to both student and teacher model for feature extraction. 
Pyramid features, \textit{e.g.} ($\{P_2, P_3, P_4, P_5\}$), are generated for both of them. 
Then, the region of interests (RoIs) generated by the student are used to crop regions in the pyramids 
for both of the student and the teacher detector. 
And RPN is followed by two fully-connected layers (FCs) 
and a pair of siblings for computing the final output classes and boxes. 
\begin{figure}[t]
\begin{center}
\includegraphics[width=1.0\linewidth]{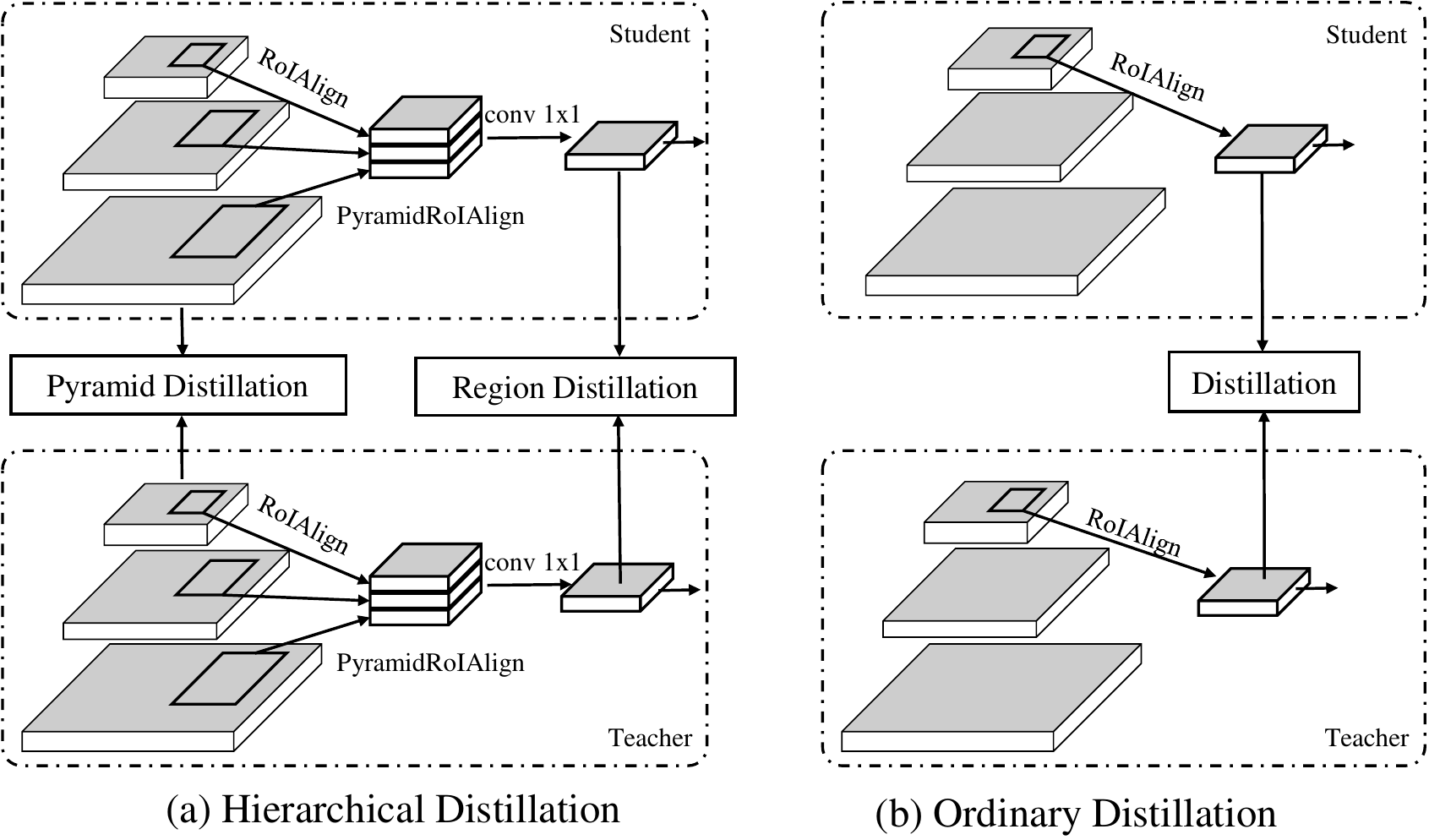}
\end{center}
\setlength{\belowcaptionskip}{-10pt} 
\vspace{-2mm}
\caption{Illustration of our proposed hierarchical knowledge distillation (a) 
versus the ordinary distillation (b).
}
\label{fig:connection}
\vspace{-4mm}
\end{figure}
\subsection{Hierarchical knowledge distillation}
Pyramid Distillation, Region Distillation and Logit Distillation 
form our comprehensive, hierarchical `teacher -- student' learning architecture. 
\subsubsection{Pyramid distillation}
Inputs are first fed into FPNs to generate feature pyramids for both student and teacher models, 
denoted by $\{P_2, P_3, P_4, P_5\}$. 
Different levels in the feature pyramid often contains information with different semantic meanings. 
The lowest level $P_2$ contains only local details, such as edges and contours. 
As we going deeper with the pyramid level, for instance, the highest level $P_5$, 
contains more abstractions or stronger semantic meanings, \textit{e.g.} object parts. 
Instead of letting the student only focuses on learning the abstractions, 
it is equivalently important for learning the details. 
Previous methods often omit this learning objective, 
to tackle above problem, 
we propose pyramid distillation for learning better object representations 
with both strong abstractions and precise details. 
As shown in Fig. 2 (a), the Pyramid Distillation is defined as: 
\begin{equation}
  L_{PD}(P^{s}, P^{t}) = \frac{1}{N_{PD}}\sum_{i=2}^{5}\left\|P_i^{s} - P_i^{t}\right\|_2,
\end{equation}
where $N_{PD}$ is the number of total locations in the pyramid, 
$P_i^s$, $P_i^t$ represents the $i$-th level of student's feature pyramid 
and that of teacher's feature pyramid, respectively. 

\subsubsection{Region distillation}
RoIAlign is first introduced in Mask R-CNN \cite{he2017mask}. 
A certain level of feature in the feature pyramid is first found according to the area of each RoI, 
then the region feature is cropped within that level, as shown in Fig. 2 (b). 
In this work, for each RoI we crop features from all levels of the feature pyramid, 
as shown in Fig. 2 (a). 
The resulting region feature ($R^{\cdot}$) contains both low-level details and high-level abstractions. 
Based on Pyramid RoIAlign, we define Region Distillation as: 
\begin{equation}
  L_{RD}(R^{s}, R^{t}) = \frac{1}{N_R}\left\|R^{s} - R^{t}\right\|_2,
\end{equation}
where $N_{R}$ is the number of total locations in the regions. 
\subsubsection{Logit distillation}

Additionally, we also add a final distillation in the training stage. In the second stage of a detector, we add 
Logit Distillation on the fully connected layer before the detection output, 
which forces the student to mimic the teacher's final behavior. 
It is defined as: 
\begin{equation}
  L_{LD}(G^{s}, G^{t}) = \frac{1}{N_G}\left\|G^{s} - G^{t}\right\|_2,
\end{equation}
where $N_{G}$ is number of proposals in the second stage detector. 

\subsubsection{Final hierarchical distillation objective} 
The final objective of our proposed hierarchical knowledge distillation is: 
\begin{equation}
  L_{dist} = \lambda_{PD}L_{PD} + \lambda_{RD}L_{RD} + \lambda_{LD}L_{LD},
\end{equation}
where $\lambda_{PD},  \lambda_{RD},  \lambda_{LD}$ are factors, which balance 
all the objectives so that they are at the same magnitude. 
\section{Experiments}
\label{sec:experiment}
%
In this section, 
we first introduce dataset, evaluation metrics, implementation details and overall performance. 
Then, we perform ablation study to validate the contributing factors proposed in this work. 
At the end, we compare our unified framework with other state-of-the-art methods. 
%

\textbf{Dataset.} 
We evaluate our unified, hierarchical knowledge distillation framework on Caltech dataset \cite{Dollar2012PAMI}. 
This pedestrian detection dataset contains 2.5 hours videos captured from a moving vehicle. 
We train our model on Caltech10$\times$, 
which samples ten times frames from the videos than original training set \cite{zhang2016far}, 
but using the original annotations still. 
In evaluation, 
we test the model on the standard test set and report results on \textit{Reasonable} configuration. 

\textbf{Evaluation Metrics.} The MR (log-average miss rate) 
between $[10^{-2}, 10^0]$ FPPI (false positive per image) 
is used as the evaluation metrics following Dollar \emph{et al.}~\cite{Dollar2012PAMI}. 
We report MRs on different subsets, 
including the \textit{reasonable} subset ($height > 50$, $visibility > 0.65$), 
and the \textit{small} subset ($75 > height > 50$, $0.65 > visibility > 0.20$). 

\textbf{Implementation Details.} Our model is trained for 6 epochs using a SGD optimizer with initial learning rate of 0.002. 
And we decrease the learning rate by a factor of 0.1 at the 4th and 6th epochs. 
The size of input image is set to $960 \times 720$, which is 1.5 times of the original image. 
We adopt randomly horizontal flipping as the only data augmentation. 
In the final distillation objective, 
the parameters $\lambda_{PD}$, $\lambda_{RD}$ and $\lambda_{LD}$ is set to 
0.5, 30 and 30, respectively. 

\textbf{Overall performance.} In Table 1, 
training through our proposed distillation framework, 
with 6 times compression in number of parameters, 
our student model still achieves competitive performance as the teacher model, 
with 10.03\% MR on \textit{reasonable} subset and 12.28\% MR on \textit{small} subset. 
%
\vspace{-3mm}
\begin{table}[h]
\setlength{\belowcaptionskip}{-1pt} 
\caption{Comparison between student and teacher model. 
Performance is measured by MR (in \%), 
lower value is better. 
}
\vspace{-2mm}
\small
\centering
\begin{threeparttable}
\label{table:reid}
\begin{tabular}{@{}l | c | c | c}
\toprule
Method & Parameters & Reasonable & Small \\
\midrule
Teacher & 68M & 9.29 & 11.86 \\ 
Student (baseline) & 11M & 12.52 & 15.96 \\ 
Student (best) & \textbf{11M} & \textbf{10.03} & \textbf{12.28} \\
\bottomrule
\end{tabular}
\end{threeparttable}
\vspace{-4mm}
\end{table}
\vspace{-3mm}
\begin{table}[h]
\setlength{\belowcaptionskip}{-1pt} 
\caption{Comparison of multiple intermediate supervision. 
}
\vspace{-2mm}
\small
\centering
\begin{threeparttable}
\label{table:reid}
\begin{tabular}{ @{}c | c c c c | c | c}
\toprule
Num & PD & RD & LD & PyRoIAlign & Reasonable & Small \\
\midrule
1 & - & - & - & - & 12.85 & 16.62 \\
2 & - & - & - & \checkmark & \textbf{12.52} & \textbf{15.96} \\ 
3 & - & - & \checkmark & \checkmark & 11.43 & 14.96 \\ 
4 & - & \checkmark & - & - & 11.95 & 14.61 \\ 
5 & - & \checkmark & - & \checkmark & 10.82 & 13.04 \\
6 & - & \checkmark & \checkmark & \checkmark & 10.74 & 13.18 \\ 
7 & \checkmark & \checkmark & \checkmark & - & 11.83 & 14.59 \\ 
8 & \checkmark & \checkmark & \checkmark & \checkmark & \textbf{10.03} & \textbf{12.28} \\
\bottomrule
\end{tabular}
\end{threeparttable}
\vspace{-4mm}
\end{table}
\subsection{Ablation study}
%
In Table 2, we validate the contributing factors proposed in this work, 
including Pyramids Distillation (PD), Region Distillation (RD), 
Logit Distillation (LD) and Pyramid RoIAlign (PyRoIAlign). 
Compared with the \textit{baseline} (\textit{2nd row} of Table 2), 
using LD (\textit{3rd row}) as the only intermediate supervision improves our performance by 1.09\%. 
Only using RD (\textit{5th row}) as supervision shows more significant improvement by 1.70\%. 
It proves that region feature contains more supervision than the final output logit, 
because the dimension of logit is significantly reduced for passing through two FC layers. 
PyRoIAlign is critical to our distillation framework, 
which boosts our performance by a notable margin. 
Using RD without PyRoIAlign (\textit{4th row}) merely boosts our performance by 0.57\%, 
which drops 1.13\% compared with enabling PyRoIAlign (\textit{5th row}). 
This performance degradation is caused by 
using supervision only from single-level features for distillation, 
as mentioned earlier in \textit{Sec 3.2.1}. 
Moreover, we combine supervisions from both RD and LD at the same time (\textit{6th row}), 
which gains notable improvements by 1.78\%. 
Further more, 
we add supervision from PD to form our comprehensive, hierarchical distillation framework (\textit{8th row}), 
and experiment result demonstrates the effectiveness of proposed framework, 
with 2.49\% performance boosting compared with \textit{baseline} method. 
Comparing with experiment without PD (\textit{6th row}), 
this experiment (\textit{8th row}) also proves that PD is critical to our unified framework, 
which contributes 0.71\% performance improvements. 
Most importantly, 
this experiment (\textit{8th row}) shows the \textit{best} performance of our proposed student model, 
which achieves 10.03\% MR in the \textit{reasonable} subset. 
And it is a very competitive performance compared with our teacher model, 
thus further proves the effectiveness of our comprehensive, hierarchical distillation framework. 
%
\vspace{-3mm}
\begin{table}[h]
\setlength{\belowcaptionskip}{-1pt} 
\caption{Comparison with other state-of-the-art methods.}
\vspace{-2mm}
\small
\centering
\begin{threeparttable}
\label{table:reid}
\begin{tabular}{@{}l | c | c}
\toprule
Method & Parameters & Reasonable \\
\midrule
CompACT-Deep \cite{cai2015learning} & 138M & 11.7 \\
UDN+SS \cite{ouyang2018jointly} & 138M & 11.5 \\ 
FstrRCNN-ATT \cite{zhang2018occluded} & 138M & 10.3 \\ 
MS-CNN \cite{cai2016unified} & 138M & 9.9 \\ 
RPN-BF \cite{zhang2016faster} & 138M & 9.7 \\ 
\midrule
Teacher (Ours) & 68M & 9.3 \\ 
Student (Ours) & \textbf{11M} & \textbf{10.0} \\
\bottomrule
\end{tabular}
\end{threeparttable}
\vspace{-4mm}
\end{table}
\subsection{Comparison with other methods} 
In Table 3, we compare our proposed method with other state-of-the-art methods on Caltech dataset. 
Our student model outperforms some of state-of-the-art methods by a notable margin, 
even if they use stronger network architecture (VGG-16) than ours. 
Though our student model is much smaller in number of parameters, 
we still achieves 1.7\% lower MR than Cai \emph{et al.}~\cite{cai2015learning}, 
1.5\% lower MR than Ouyang \emph{et al.}~\cite{ouyang2018jointly}, 
and 0.3\% lower MR than Zhang \emph{et al.}~\cite{zhang2018occluded}. 
Even with limited representation ability, 
our model still achieves competitive performance 
with other state-of-the-art methods \cite{cai2016unified, zhang2016faster}. 
\section{Conclusion}
\label{sec:majhead}
We propose a comprehensive, hierarchical knowledge distillation framework 
for training lightweight pedestrian detector. 
Comparing with single-level supervision, 
our unified framework utilize multiple intermediate supervisions for distillation, 
which significantly improves the efficiency of transferring 
knowledge from a teacher model to a student model. 
Besides, our hierarchical distillation framework helps our student model 
learn better representations from multi-level feature maps 
with both abstractions and details. 
Experiment results on Caltech pedestrian detection benchmark 
demonstrate the effectiveness of our proposed framework. 

\bibliographystyle{IEEEbib}
\bibliography{refs}

\begin{thebibliography}{10}

\bibitem{krizhevsky2012imagenet}
Alex Krizhevsky, Ilya Sutskever, and Geoffrey~E Hinton,
\newblock ``Imagenet classification with deep convolutional neural networks,''
\newblock in {\em NeurIPS}, 2012.

\bibitem{Simonyan15}
K.~Simonyan and A.~Zisserman,
\newblock ``Very deep convolutional networks for large-scale image
  recognition,''
\newblock in {\em ICLR}, 2015.

\bibitem{he2016deep}
Kaiming He, Xiangyu Zhang, Shaoqing Ren, and Jian Sun,
\newblock ``Deep residual learning for image recognition,''
\newblock in {\em CVPR}. IEEE, 2016.

\bibitem{girshick2014rich}
Ross Girshick, Jeff Donahue, Trevor Darrell, and Jitendra Malik,
\newblock ``Rich feature hierarchies for accurate object detection and semantic
  segmentation,''
\newblock in {\em CVPR}. IEEE, 2014.

\bibitem{girshick2015fast}
Ross Girshick,
\newblock ``Fast r-cnn,''
\newblock in {\em ICCV}. IEEE, 2015.

\bibitem{ren2015faster}
Shaoqing Ren, Kaiming He, Ross Girshick, and Jian Sun,
\newblock ``Faster r-cnn: Towards real-time object detection with region
  proposal networks,''
\newblock in {\em NeurIPS}, 2015.

\bibitem{lin2017feature}
Tsung-Yi Lin, Piotr Doll{\'a}r, Ross Girshick, Kaiming He, Bharath Hariharan,
  and Serge Belongie,
\newblock ``Feature pyramid networks for object detection,''
\newblock in {\em CVPR}. IEEE, 2017.

\bibitem{hinton2015distilling}
Geoffrey Hinton, Oriol Vinyals, and Jeff Dean,
\newblock ``Distilling the knowledge in a neural network,''
\newblock {\em arXiv preprint arXiv:1503.02531}, 2015.

\bibitem{romero2014fitnets}
Adriana Romero, Nicolas Ballas, Samira~Ebrahimi Kahou, Antoine Chassang, Carlo
  Gatta, and Yoshua Bengio,
\newblock ``Fitnets: Hints for thin deep nets,''
\newblock {\em ICLR}, 2015.

\bibitem{wei2018quantization}
Yi~Wei, Xinyu Pan, Hongwei Qin, Wanli Ouyang, and Junjie Yan,
\newblock ``Quantization mimic: Towards very tiny cnn for object detection,''
\newblock in {\em ECCV}. Springer, 2018.

\bibitem{Li_2017_CVPR}
Quanquan Li, Shengying Jin, and Junjie Yan,
\newblock ``Mimicking very efficient network for object detection,''
\newblock in {\em CVPR}. IEEE, 2017.

\bibitem{chen2017learning}
Guobin Chen, Wongun Choi, Xiang Yu, Tony Han, and Manmohan Chandraker,
\newblock ``Learning efficient object detection models with knowledge
  distillation,''
\newblock in {\em NeurIPS}, 2017.

\bibitem{Dollar2012PAMI}
Piotr Doll\'ar, Christian Wojek, Bernt Schiele, and Pietro Perona,
\newblock ``Pedestrian detection: An evaluation of the state of the art,''
\newblock {\em TPAMI}, vol. 34, no. 4, pp. 743--761, 2012.

\bibitem{redmon2016you}
Joseph Redmon, Santosh Divvala, Ross Girshick, and Ali Farhadi,
\newblock ``You only look once: Unified, real-time object detection,''
\newblock in {\em CVPR}. IEEE, 2016.

\bibitem{liu2016ssd}
Wei Liu, Dragomir Anguelov, Dumitru Erhan, Christian Szegedy, Scott Reed,
  Cheng-Yang Fu, and Alexander~C Berg,
\newblock ``Ssd: Single shot multibox detector,''
\newblock in {\em ECCV}. Springer, 2016.

\bibitem{lin2018focal}
Tsung-Yi Lin, Priyal Goyal, Ross Girshick, Kaiming He, and Piotr Doll{\'a}r,
\newblock ``Focal loss for dense object detection,''
\newblock {\em TPAMI}, 2018.

\bibitem{shang2018zoomnet}
Chong Shang, Haizhou Ai, Zijie Zhuang, Long Chen, and Junliang Xing,
\newblock ``Zoomnet: Deep aggregation learning for high-performance small
  pedestrian detection,''
\newblock in {\em ACML}, 2018.

\bibitem{he2017mask}
Kaiming He, Georgia Gkioxari, Piotr Doll{\'a}r, and Ross Girshick,
\newblock ``Mask r-cnn,''
\newblock in {\em ICCV}. IEEE, 2017.

\bibitem{zhang2016far}
Shanshan Zhang, Rodrigo Benenson, Mohamed Omran, Jan Hosang, and Bernt Schiele,
\newblock ``How far are we from solving pedestrian detection?,''
\newblock in {\em CVPR}. IEEE, 2016.

\bibitem{cai2015learning}
Zhaowei Cai, Mohammad Saberian, and Nuno Vasconcelos,
\newblock ``Learning complexity-aware cascades for deep pedestrian detection,''
\newblock in {\em CVPR}. IEEE, 2015.

\bibitem{ouyang2018jointly}
Wanli Ouyang, Hui Zhou, Hongsheng Li, Quanquan Li, Junjie Yan, and Xiaogang
  Wang,
\newblock ``Jointly learning deep features, deformable parts, occlusion and
  classification for pedestrian detection,''
\newblock {\em TPAMI}, vol. 40, no. 8, pp. 1874--1887, 2018.

\bibitem{zhang2018occluded}
Shanshan Zhang, Jian Yang, and Bernt Schiele,
\newblock ``Occluded pedestrian detection through guided attention in cnns,''
\newblock in {\em CVPR}. IEEE, 2018.

\bibitem{cai2016unified}
Zhaowei Cai, Quanfu Fan, Rogerio~S Feris, and Nuno Vasconcelos,
\newblock ``A unified multi-scale deep convolutional neural network for fast
  object detection,''
\newblock in {\em ECCV}. Springer, 2016.

\bibitem{zhang2016faster}
Liliang Zhang, Liang Lin, Xiaodan Liang, and Kaiming He,
\newblock ``Is faster r-cnn doing well for pedestrian detection?,''
\newblock in {\em ECCV}. Springer, 2016.

\end{thebibliography}

\end{document}